\pdfoutput=1

\documentclass[11pt]{article}

\usepackage{ACL2023}

\usepackage{times}
\usepackage{latexsym}

\usepackage[T1]{fontenc}
\usepackage[utf8]{inputenc}
\usepackage{microtype}
\usepackage{inconsolata}

\usepackage{graphicx}
\usepackage{CJKutf8}
\usepackage{booktabs}
\usepackage{bm}
\usepackage{amsmath}
\usepackage{multirow}
\usepackage{subfigure}

\title{Towards Speech Dialogue Translation Mediating Speakers\\of Different Languages}

\author{
  Shuichiro Shimizu${}^{1}$ \, Chenhui Chu${}^{1}$ \, Sheng Li${}^{2}$ \, Sadao Kurohashi${}^{1,3}$ \\
${}^{1}$Kyoto University, Japan \\
${}^{2}$National Institute of Information and Communications Technology, Japan \\
${}^{3}$National Institute of Informatics, Japan \\
\texttt{\{sshimizu,chu,kuro\}@nlp.ist.i.kyoto-u.ac.jp} \, \texttt{sheng.li@nict.go.jp}
}

\begin{document}
\maketitle
\begin{abstract}
We present a new task, speech dialogue translation mediating speakers of different languages.
We construct the SpeechBSD dataset for the task and conduct baseline experiments.
Furthermore, we consider context to be an important aspect that needs to be addressed in this task and propose two ways of utilizing context, namely monolingual context and bilingual context.
We conduct cascaded speech translation experiments using Whisper and mBART, and show that bilingual context performs better in our settings.
\end{abstract}

\section{Introduction}
In this global era, it is becoming increasingly important for people from different countries/regions to interact with each other and have a mutual understanding.
Recent advancements in machine translation (MT) technologies have enabled us to communicate with people worldwide, especially in text.
Chat translation or dialogue machine translation \citep{liu-etal-2021-dialoguemt} supports such communications, which enables people who use different languages to have cross-language chats.
Speech translation (ST) has also recently shown success (e.g., \citealp{chen-etal-2022-maestro}), especially in monologue translation (e.g., \citealp{di-gangi-etal-2019-mustc}).
However, to the best of our knowledge, no study has focused on ST of dialogues, which is an important aspect of language usage.

In this study, we propose a new task: speech dialogue translation (SDT) aiming to mediate speakers of different languages.
We consider bilingual dialogues where several people who speak in different languages talk with each other mediated by an ST system.

\begin{figure*}[t]
    \centering
    \includegraphics[width=1.5\columnwidth]{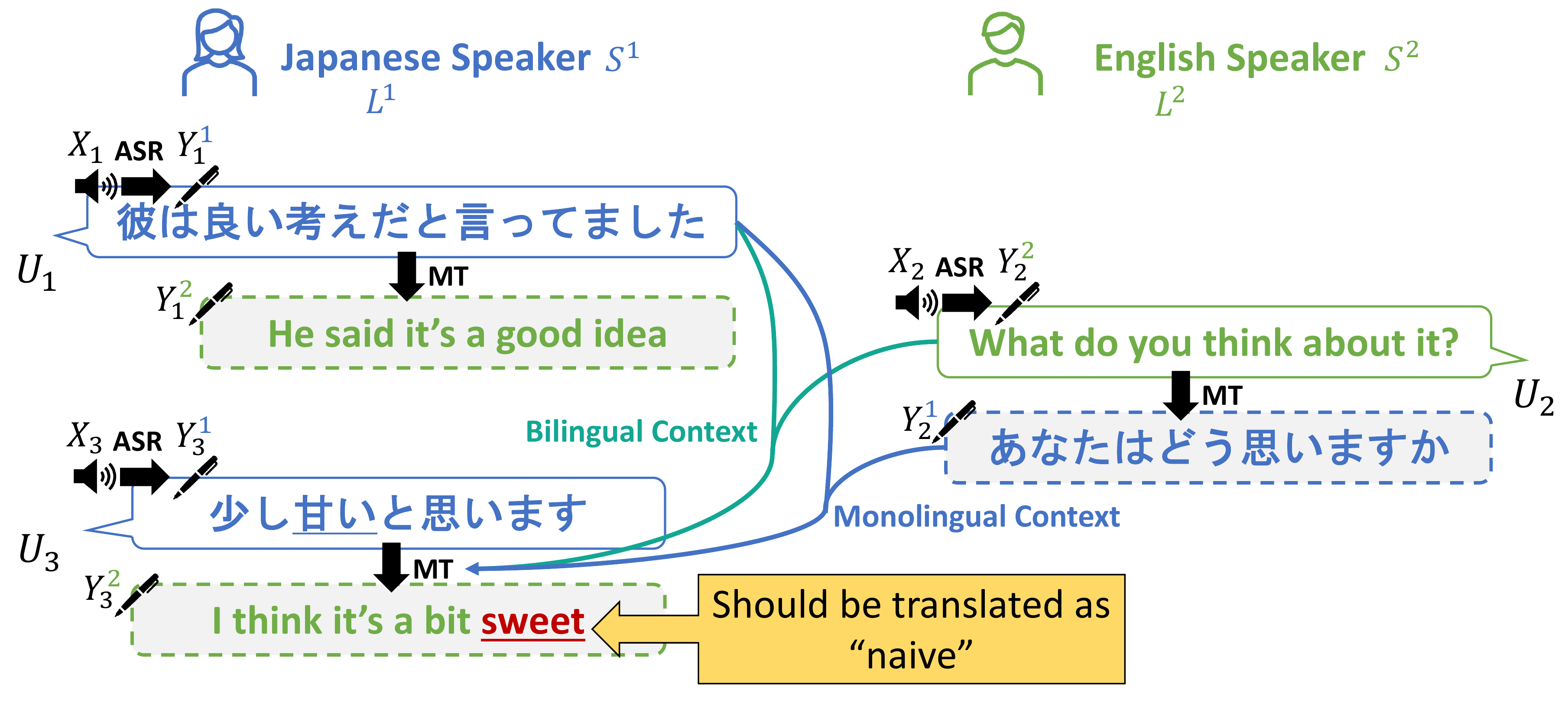}
    \caption{The importance of considering context in SDT. ``\begin{CJK}{UTF8}{ipxm}甘い\end{CJK}'' can be translated into either ``sweet'' or ``naive,'' which can be disambiguated with the context. We consider two types of context for translation, monolingual context and bilingual context.}
    \label{fig:sdt}
\end{figure*}

It is important to consider context in SDT because we need to consider context in different languages, which cannot be readily handled by current ST systems that mainly focus on one translation direction.
Figure \ref{fig:sdt} shows an example of an ST-mediated dialogue between an English speaker and a Japanese speaker.
They are discussing some ideas, and the English speaker says, ``What do you think about it?''
The Japanese speaker responds by saying the idea is naive, but without context it can be translated as ``I think it's a bit sweet'' because ``\begin{CJK}{UTF8}{ipxm}甘い\end{CJK}'' has two meanings, sweet and naive.
By utilizing dialogue context, the meaning of ``\begin{CJK}{UTF8}{ipxm}甘い\end{CJK}'' becomes clear so that the utterance can be translated properly.

For the proposed task, we construct the SpeechBSD dataset\footnote{The dataset is made public under the CC BY-NC-SA 4.0 license at \url{https://github.com/ku-nlp/speechBSD}.} based on an existing text dialogue corpus, BSD (Bussiness Scene Dialogue) corpus \citep{rikters-etal-2019-bsd}.
We collect audio of the BSD corpus through crowdsourcing along with speaker attributes.

We conduct speech-to-text cascaded ST experiments on the dataset.
There are two mainstream methods for ST, the cascade method \citep{stentiford-and-steer-cascade-1988} where automatic speech recognition (ASR) and MT are chained together, and the end-to-end method \citep{duong-etal-2016-attentional,berard-etal-2016-lat}, where translations are directly predicted from speech.
Recent study \citep{bentivogli-etal-2021-cascade,tran-etal-2022-really} suggests that the two methods are on par.
We conduct cascade ST experiments using Whisper \citep{radford-etal-2022-whisper} for ASR and mBART \citep{liu-etal-2020-mbart} for MT.

We consider three settings for translation: without context, with monolingual context, and with bilingual context.
The monolingual context is composed in the language the utterance to be translated is spoken, whereas the bilingual context is composed in the original language of the spoken utterances (see examples in Figure \ref{fig:sdt}).
We show that translation with bilingual context performs better compared to the one without context by up to $1.9$ BLEU points in MT and $1.7$ BLEU points in cascade ST with our settings.
We also conduct a manual evaluation focusing on zero anaphora, a grammatical phenomenon where arguments of verbs are omitted when they are apparent from the context in Japanese.
We show that with bilingual context, the MT models can often predict zero pronouns correctly.

\section{Related Work}

Although neural MT has greatly improved over the past few years, the translation of dialogues remains a challenging task because of its characteristics.
\citet{liu-etal-2021-dialoguemt} summarizes the recent progress of dialogue MT and categorizes its issue into four categories, coherence, consistency, cohesion, and personality.
The main approaches to address these problems include document MT (e.g., \citealp{liu-etal-2021-dialoguemt}), usage of pretrained models (e.g., \citealp{wang-etal-2020-wmt20tencent}), and auxiliary task learning utilizing speaker information (e.g., \citealp{liang-etal-2021-nct}).

Considering context in ST is recently studied for the end-to-end approach \citep{zhang-etal-2021-context}.
We point out that although not addressed in this work, considering context for ASR is also an active research area (e.g., \citealp{inaguma-kawahara-2021-unsegmentedasr}).

In this work, we focus on the translation of speech dialogue.
We use mBART, which performed best in a previous work of chat translation \citep{liu-etal-2021-dialoguemt}, and also consider utilizing context.

\section{Speech Dialogue Translation (SDT)}
In SDT, there are several speakers who speak different languages with the help of a translation system.
In this work, we consider $M$ speakers $\{ S^m \, | \, m=1, 2, \cdots, M \}$ and 2 languages $\{ L^n \, | \, n=1, 2 \}$.
We consider a dialogue with $T$ utterances $D = (U_1, \cdots, U_T)$, where an utterance is $U_t = (S^m_t, L^n_t, X_t)$.
Here, $S_t^m$ is the speaker, $L_t^n$ is the language spoken, and $X_t$ is the speech signal of $t$-th utterance.
Let $Y_t^n \, (n=1, 2)$ be text that has the same meaning as $X_t$ in language $L^n$.
The task of SDT is to generate translation $Y_t^2$ from speech signal $X_t$ when the source language is $L^1$ (or translation $Y_t^1$ from $X_t$ when the source language is $L^2$) for every utterance $U_t$.

\section{SpeechBSD Dataset}
We construct the SpeechBSD dataset to study SDT.
It is based on the existing dialogue dataset in text, BSD corpus \citep{rikters-etal-2019-bsd,rikters-etal-2021-bsdjournal}.
We collect audio of all the sentences in the dataset along with speaker attributes (gender and homeplace) through crowdsourcing.

\subsection{BSD Corpus}
BSD corpus is a parallel corpus of English and Japanese composed of manually designed business scene dialogues.
Each dialogue called scenario contains $30$ sentences on average spoken by $2$-$5$ speakers.
The original language the scenarios were written in is half English and half Japanese so that the expressions are not biased toward one language.

\subsection{Dataset Construction}
First, we divided each scenario by speaker.
For example in Figure \ref{fig:sdt}, the original BSD corpus contains text of $Y_1^1, Y_1^2, Y_2^1, Y_2^2, Y_3^1$, and $Y_3^2$.
In this case, we divide the dialogue into four parts: the Japanese speaker part ($Y_1^1$ and $Y_3^1$), the English speaker part ($Y_2^2$), another Japanese speaker part ($Y_2^1$), and another English speaker part ($Y_1^2$ and $Y_3^2$).
In this way, we can compose two cross-language dialogues ($Y_1^1 \rightarrow Y_2^2 \rightarrow Y_3^1$ and $Y_1^2 \rightarrow Y_2^1 \rightarrow Y_3^2$) from one scenario of the BSD corpus.
We collected audio through crowdsourcing so that each part is spoken by a different worker.\footnote{Because workers could participate in multiple tasks, it is possible that different parts are actually spoken by the same person.}
We designed a web application to record audio and collected English speech from the US using Amazon Mechanical Turk\footnote{\url{https://www.mturk.com/}} and Japanese speech from Japan using Yahoo! crowdsourcing.\footnote{\url{https://crowdsourcing.yahoo.co.jp/}}
We also collected the gender and homeplace (the US states or Japanese prefecture) of the speakers as they may affect translation performance.
The instructions given to the workers are shown in Appendix \ref{sec:crowdsourcing-instruction}.

\subsection{Statistics of the SpeechBSD Dataset}
The collected audio was $24.3$ hours for English speech and $30.7$ hours for Japanese speech in total.
Details are provided in Appendix \ref{sec:appendix-speechbsd-statistics} Table \ref{tab:speechbsd}.
Regarding speaker gender, English speech was balanced, whereas there were more male speakers in Japanese.
As for homeplace, in Japanese, the speakers were distributed roughly according to the population distribution.
In English, it was less diverse (Appendix \ref{sec:appendix-speechbsd-statistics} Figure \ref{fig:distribution}).

\section{Considering Context for SDT} \label{sec:context-sdt}
We propose two ways to consider context in SDT: monolingual context and bilingual context.

First, for every utterance $U_t$, an ASR system is used to obtain transcripts $Y_t^n$.

The monolingual context is composed in the source language of the utterance to be translated.
For example, in Figure \ref{fig:sdt}, when translating the third utterance $U_3$ from Japanese to English, as the source language of the utterance is Japanese ($L^1$), the context ($Y_1^1$ and $Y_2^1$) is also composed in Japanese.
Let the context composed in this way be $\bm{Y}^n_{<t}$.

For monolingual context experiments, we use two translation models for each translation direction.
The training objective of the MT model that translates from $L^1$ to $L^2$ is to maximize the following log likelihood\footnote{The utterances are generated token-wise. The notations in equations \ref{eq:monocontext-objective} and \ref{eq:bicontext-objective} are simplified for clarity.}:
\begin{equation}
    \mathcal{L}^{1\rightarrow2} = \sum_t \log \mathrm{P} (Y_t^2, \bm{Y}^2_{<t} \, | \, Y_t^1, \bm{Y}^1_{<t}).  \label{eq:monocontext-objective}
\end{equation}
Similar objective $\mathcal{L}^{2\rightarrow1}$ can be derived when $L^2$ is the source language and $L^1$ is the target language.
Postprocessing is applied to extract $Y_t^2$ from the output that contains both $\bm{Y}^2_{<t}$ and $Y_t^2$.

The bilingual context is composed of the original language of the spoken utterances.
For example, in Figure \ref{fig:sdt}, when translating the third utterance $U_3$ from Japanese to English, the bilingual context on the source side is $Y_1^1$ and $Y_2^2$, which involves both languages.
The bilingual context on the target side is $Y_1^2$ and $Y_2^1$.
Because there is no concept of source or target language in this case, let the source side utterance be $Y_t$, source side context be $\bm{Y}_{<t}$, target side utterance be $\overline{Y_t}$, and target side context be $\overline{\bm{Y}_{<t}}$.
The MT model is trained with the following objective:
\begin{equation}
    \mathcal{L} = \sum_t \log \mathrm{P} (\overline{Y_t}, \overline{\bm{Y}_{<t}} \, | \, Y_t, \bm{Y}_{<t}).  \label{eq:bicontext-objective}
\end{equation}
Postprocessing is applied to extract $\overline{Y_t}$ from the output.

We consider constrained context with context size $c$ in practice, which shows the number of previous utterances used for translation in addition to the utterance to be translated.
More formal definitions of monolingual, bilingual, and constrained context are provided in Appendix \ref{sec:formal-definition}.

\section{Experiments}
\subsection{Automatic Speech Recognition}
In SDT, ASR has to handle bilingual inputs.
We used a multilingual ASR model Whisper \citep{radford-etal-2022-whisper}.
The medium model with 12 encoder and decoder layers was used without finetuning.
Further details are provided in Appendix \ref{sec:asr-settings}.
We evaluated the performance of the SpeechBSD test set.
For English the word error rate was $8.3$ \%, and for Japanese   the character error rate was $13.2$ \%.

\subsection{Machine Translation} \label{sec:mt}
MT model also needs to handle bilingual inputs in SDT.
We used mBART \citep{liu-etal-2020-mbart} and finetuned the model with SpeechBSD for MT.
The large model with 12 encoder and decoder layers was used.
Although the dialogues are regarded as bilingual ones in this study, the predictions were recomposed to the monolingual dialogue form for evaluation because usually performance of MT models is evaluated on a single language pair.
SacreBLEU \citep{post-2018-sacrebleu} was used for calculating BLEU scores.
Further details are provided in Appendix \ref{sec:mt-settings}.

\subsubsection{Context Settings}
Three settings were considered: translation without context, with monolingual context, and with bilingual context.

\paragraph{Without Context}
Each utterance in a scenario was treated as a separate sentence in this setting.
Finetuning was performed separately for each translation direction.

\paragraph{Monolingual Context}
For each utterance in a scenario, monolingual context with context width $c = 5$ was composed in the way described in section \ref{sec:context-sdt}.
The context utterances and the utterance to translate were concatenated with the end of sentence token \texttt{</s>}.
Finetuning was performed separately for each translation direction.

\paragraph{Bilingual Context}
For each utterance in a scenario, bilingual context with context width $c = 5$ was composed in the way described in section \ref{sec:context-sdt}.
The context utterances and the utterance to translate were concatenated with the end of sentence token \texttt{</s>}.
As there is no concept of source language or target language in this setting, a single model was finetuned in this setting.

\subsubsection{Results}
Table \ref{tab:mt-and-cascade} (upper part) shows the results of the MT experiments.
Comparing ``Without'' with ``Monolingual,'' more than $0.9$ points of improvement were observed using monolingual context.
Comparing ``Monolingual'' with ``Bilingual,'' the latter performed better, especially in Ja--En.

\begin{table}[t]
    \centering
    \small
    \begin{tabular}{@{}ccll@{}}
    \toprule
     & Context & En--Ja & Ja--En \\ \midrule
    \multirow{3}{*}{MT} & Without & 15.9 & 18.2 \\
     & Monolingual & 16.8$^\dag$  & 19.5$^\dag$ \\
     & Bilingual & \textbf{17.0}$^\dag$  & \textbf{20.1}$^{\dag\ddag}$ \\ \midrule

    \multirow{3}{*}{Cascade ST} & Without & 15.2 & 15.4 \\
     & Monolingual & 15.9$^\dag$  & 16.5$^\dag$ \\
     & Bilingual & \textbf{16.4}$^\dag$  & \textbf{17.1}$^{\dag\ddag}$ \\ \bottomrule
    \end{tabular}
    \caption{BLEU scores of the SpeechBSD test set for MT and Cascade ST experiments. 
    ``$\dag$'' and ``$\ddag$'' indicate that the results are significantly better than ``without context'' and ``monolingual context'' at $p < 0.05$, respectively.
    }
    \label{tab:mt-and-cascade}
\end{table}

\subsubsection{Manual Evaluation} \label{sec:manual-eval}
To verify how context can help improve translations, we conducted a manual evaluation focusing on a grammatical phenomenon called zero anaphora, as discussed in \citet{rikters-etal-2019-bsd}.
Similarly to \citet{rikters-etal-2019-bsd}, we counted the number of sentences with pronouns \textit{I, you, he, she, it}, and \textit{they} in English\footnote{We tokenized sentences with the NLTK toolkit \cite{bird-etal-2009-nltk}.} and observed that $63$ \% of the test sentences included them.
We sampled $50$ of those sentences from the test set.
First, we checked if the subjects of the Japanese sentences were zero pronouns by comparing Japanese and English gold references.
Then we checked if the zero pronouns were translated into English correctly for the predictions of each Ja--En system.

Out of the $50$ sentences, $29$ were sentences with zero pronoun subjects.
The number of sentences that the missing pronoun was translated correctly was $19, 20$, and $24$ for without context, monolingual context, and bilingual context settings, respectively.
This shows that context can help disambiguate zero pronouns, and using bilingual context can help generate correct pronouns.
Examples of the sentences are shown in Appendix \ref{sec:manual-eval-example}.

\subsection{Cascade Speech Translation}
Cascade ST experiments were performed by using Whisper recognition results as input to the MT models described in section \ref{sec:mt}.

Table \ref{tab:mt-and-cascade} (lower part) shows the results.
Similarly to MT, BLEU score improved by more than $0.7$ points by using monolingual context.
Further improvements by more than $0.5$ points were observed using bilingual context.

We also performed manual evaluation as in Section \ref{sec:manual-eval}.
The number of sentences that the missing pronoun was translated correctly was $16$, $18$, and $22$ for without context, monolingual context, and bilingual context settings, respectively.
It showed a similar trend to the results of section \ref{sec:manual-eval} with lower translation accuracy.
Examples of the sentences are shown in Appendix \ref{sec:manual-eval-example}.

\section{Conclusion}
We presented a new task, SDT aiming to mediate speakers of different languages.
We constructed the SpeechBSD dataset via crowdsourcing.
We performed MT experiments utilizing context and showed its effectiveness.
In the future, we plan to perform experiments in end-to-end ST settings and SDT utilizing speaker attributes.

\section*{Limitations}
The experiments were performed only on Japanese and English bilingual dialogue collected from a limited number of native speakers.
Although the methods proposed in this work can work on any language pair, drawing conclusions for other language pairs should be avoided.
The experiments were performed using existing pretrained models, Whisper and mBART, and the method used to pretrain those models would have affected the translation performances in this work.
The dialogues in the SpeechBSD dataset are the read speech of pre-composed text dialogues, and further research is required for more realistic settings such as spontaneous dialogues.

\section*{Ethics Statement}
Consent was obtained from the crowdsourcing workers when collecting audio, gender, and homeplace.
The SpeechBSD dataset is made public under the Creative Commons Attribution-NonCommercial-ShareAlike (CC BY-NC-SA) 4.0 license, which is the same as the license of the BSD corpus, and shall be used only for research purposes.
Caution should be exercised when using gender or homeplace information included in the dataset so that the identities of the speakers are not revealed.

\section*{Acknowledgegments}
This work was supported by JSPS KAKENHI Grant Numbers JP23H03454 and JP23KJ1356.

\bibliography{custom}

\begin{thebibliography}{21}
\expandafter\ifx\csname natexlab\endcsname\relax\def\natexlab#1{#1}\fi

\bibitem[{Bentivogli et~al.(2021)Bentivogli, Cettolo, Gaido, Karakanta,
  Martinelli, Negri, and Turchi}]{bentivogli-etal-2021-cascade}
Luisa Bentivogli, Mauro Cettolo, Marco Gaido, Alina Karakanta, Alberto
  Martinelli, Matteo Negri, and Marco Turchi. 2021.
\newblock \href {https://doi.org/10.18653/v1/2021.acl-long.224} {Cascade versus
  direct speech translation: Do the differences still make a difference?}
\newblock In \emph{Proceedings of the 59th Annual Meeting of the Association
  for Computational Linguistics and the 11th International Joint Conference on
  Natural Language Processing (Volume 1: Long Papers)}, pages 2873--2887,
  Online. Association for Computational Linguistics.

\bibitem[{Berard et~al.(2016)Berard, Pietquin, Servan, and
  Besacier}]{berard-etal-2016-lat}
Alexandre Berard, Olivier Pietquin, Christophe Servan, and Laurent Besacier.
  2016.
\newblock Listen and {T}ranslate: {A} {P}roof of {C}oncept for {E}nd-to-{E}nd
  {S}peech-to-{T}ext {T}ranslation.
\newblock In \emph{NIPS Workshop on End-to-end Learning for Speech and Audio
  Processing}.

\bibitem[{Bird et~al.(2009)Bird, Loper, and Klein}]{bird-etal-2009-nltk}
Steven Bird, Edward Loper, and Ewan Klein. 2009.
\newblock \emph{{Natural Language Processing with Python}}.
\newblock O'Reilly Media Inc.

\bibitem[{Chen et~al.(2022)Chen, Zhang, Rosenberg, Ramabhadran, Moreno, Bapna,
  and Zen}]{chen-etal-2022-maestro}
Zhehuai Chen, Yu~Zhang, Andrew Rosenberg, Bhuvana Ramabhadran, Pedro~J. Moreno,
  Ankur Bapna, and Heiga Zen. 2022.
\newblock \href {https://doi.org/10.21437/Interspeech.2022-10937} {{MAESTRO:
  Matched Speech Text Representations through Modality Matching}}.
\newblock In \emph{Proc. Interspeech 2022}, pages 4093--4097.

\bibitem[{Di~Gangi et~al.(2019)Di~Gangi, Cattoni, Bentivogli, Negri, and
  Turchi}]{di-gangi-etal-2019-mustc}
Mattia~A. Di~Gangi, Roldano Cattoni, Luisa Bentivogli, Matteo Negri, and Marco
  Turchi. 2019.
\newblock \href {https://doi.org/10.18653/v1/N19-1202} {{M}u{ST}-{C}: a
  {M}ultilingual {S}peech {T}ranslation {C}orpus}.
\newblock In \emph{Proceedings of the 2019 Conference of the North {A}merican
  Chapter of the Association for Computational Linguistics: Human Language
  Technologies, Volume 1 (Long and Short Papers)}, pages 2012--2017,
  Minneapolis, Minnesota. Association for Computational Linguistics.

\bibitem[{Duong et~al.(2016)Duong, Anastasopoulos, Chiang, Bird, and
  Cohn}]{duong-etal-2016-attentional}
Long Duong, Antonios Anastasopoulos, David Chiang, Steven Bird, and Trevor
  Cohn. 2016.
\newblock An {A}ttentional {M}odel for {S}peech {T}ranslation {W}ithout
  {T}ranscription.
\newblock In \emph{Proceedings of the 2016 Conference of the North {A}merican
  Chapter of the Association for Computational Linguistics: Human Language
  Technologies}.

\bibitem[{Inaguma and Kawahara(2021)}]{inaguma-kawahara-2021-unsegmentedasr}
Hirofumi Inaguma and Tatsuya Kawahara. 2021.
\newblock \href {https://doi.org/10.21437/Interspeech.2021-1107} {{VAD-Free
  Streaming Hybrid CTC/Attention ASR for Unsegmented Recording}}.
\newblock In \emph{Proc. Interspeech 2021}, pages 4049--4053.

\bibitem[{Kudo and Richardson(2018)}]{kudo-richardson-2018-sentencepiece}
Taku Kudo and John Richardson. 2018.
\newblock \href {https://doi.org/10.18653/v1/D18-2012} {{S}entence{P}iece: A
  simple and language independent subword tokenizer and detokenizer for neural
  text processing}.
\newblock In \emph{Proceedings of the 2018 Conference on Empirical Methods in
  Natural Language Processing: System Demonstrations}, pages 66--71, Brussels,
  Belgium. Association for Computational Linguistics.

\bibitem[{Liang et~al.(2021)Liang, Zhou, Meng, Xu, Chen, Su, and
  Zhou}]{liang-etal-2021-nct}
Yunlong Liang, Chulun Zhou, Fandong Meng, Jinan Xu, Yufeng Chen, Jinsong Su,
  and Jie Zhou. 2021.
\newblock \href {https://aclanthology.org/2021.emnlp-main.6} {Towards making
  the most of dialogue characteristics for neural chat translation}.
\newblock In \emph{Proceedings of the 2021 Conference on Empirical Methods in
  Natural Language Processing}, pages 67--79, Online and Punta Cana, Dominican
  Republic. Association for Computational Linguistics.

\bibitem[{Liu et~al.(2021)Liu, Sun, and Wang}]{liu-etal-2021-dialoguemt}
Siyou Liu, Yuqi Sun, and Longyue Wang. 2021.
\newblock \href {https://doi.org/10.3390/info12110484} {Recent advances in
  dialogue machine translation}.
\newblock \emph{Information}, 12(11).

\bibitem[{Liu et~al.(2020)Liu, Gu, Goyal, Li, Edunov, Ghazvininejad, Lewis, and
  Zettlemoyer}]{liu-etal-2020-mbart}
Yinhan Liu, Jiatao Gu, Naman Goyal, Xian Li, Sergey Edunov, Marjan
  Ghazvininejad, Mike Lewis, and Luke Zettlemoyer. 2020.
\newblock \href {https://doi.org/10.1162/tacl_a_00343} {Multilingual denoising
  pre-training for neural machine translation}.
\newblock \emph{Transactions of the Association for Computational Linguistics},
  8:726--742.

\bibitem[{Ott et~al.(2019)Ott, Edunov, Baevski, Fan, Gross, Ng, Grangier, and
  Auli}]{ott-etal-2019-fairseq}
Myle Ott, Sergey Edunov, Alexei Baevski, Angela Fan, Sam Gross, Nathan Ng,
  David Grangier, and Michael Auli. 2019.
\newblock \href {https://doi.org/10.18653/v1/N19-4009} {fairseq: A fast,
  extensible toolkit for sequence modeling}.
\newblock In \emph{Proceedings of the 2019 Conference of the North {A}merican
  Chapter of the Association for Computational Linguistics (Demonstrations)},
  pages 48--53, Minneapolis, Minnesota. Association for Computational
  Linguistics.

\bibitem[{Post(2018)}]{post-2018-sacrebleu}
Matt Post. 2018.
\newblock \href {https://doi.org/10.18653/v1/W18-6319} {A call for clarity in
  reporting {BLEU} scores}.
\newblock In \emph{Proceedings of the Third Conference on Machine Translation:
  Research Papers}, pages 186--191, Brussels, Belgium. Association for
  Computational Linguistics.

\bibitem[{Radford et~al.(2022)Radford, Kim, Xu, Brockman, McLeavey, and
  Sutskever}]{radford-etal-2022-whisper}
Alec Radford, Jong~Wook Kim, Tao Xu, Greg Brockman, Chiristine McLeavey, and
  Ilya Sutskever. 2022.
\newblock \href {https://cdn.openai.com/papers/whisper.pdf} {{Robust Speech
  Recognition via Large-Scale Weak Supervision}}.

\bibitem[{Riezler and Maxwell(2005)}]{riezler-maxwell-2005-pitfalls}
Stefan Riezler and John~T. Maxwell. 2005.
\newblock \href {https://aclanthology.org/W05-0908} {On some pitfalls in
  automatic evaluation and significance testing for {MT}}.
\newblock In \emph{Proceedings of the {ACL} Workshop on Intrinsic and Extrinsic
  Evaluation Measures for Machine Translation and/or Summarization}, pages
  57--64, Ann Arbor, Michigan. Association for Computational Linguistics.

\bibitem[{Rikters et~al.(2019)Rikters, Ri, Li, and
  Nakazawa}]{rikters-etal-2019-bsd}
Mat{\=\i}ss Rikters, Ryokan Ri, Tong Li, and Toshiaki Nakazawa. 2019.
\newblock \href {https://doi.org/10.18653/v1/D19-5204} {Designing the business
  conversation corpus}.
\newblock In \emph{Proceedings of the 6th Workshop on Asian Translation}, pages
  54--61, Hong Kong, China. Association for Computational Linguistics.

\bibitem[{Rikters et~al.(2021)Rikters, Ri, Li, and
  Nakazawa}]{rikters-etal-2021-bsdjournal}
Matīss Rikters, Ryokan Ri, Tong Li, and Toshiaki Nakazawa. 2021.
\newblock \href {https://doi.org/10.5715/jnlp.28.380} {Japanese–english
  conversation parallel corpus for promoting context-aware machine translation
  research}.
\newblock \emph{Journal of Natural Language Processing}, 28(2):380--403.

\bibitem[{Stentiford and Steer(1988)}]{stentiford-and-steer-cascade-1988}
F.~W.~M. Stentiford and M.~G. Steer. 1988.
\newblock Machine translation of speech.
\newblock \emph{British Telecom technology journal}.

\bibitem[{Tran et~al.(2022)Tran, Thulke, Gao, Herold, and
  Ney}]{tran-etal-2022-really}
Viet Anh~Khoa Tran, David Thulke, Yingbo Gao, Christian Herold, and Hermann
  Ney. 2022.
\newblock \href {https://arxiv.org/pdf/2210.13700.pdf} {{Does Joint Training
  Really Help Cascaded Speech Translation?}}

\bibitem[{Wang et~al.(2020)Wang, Tu, Wang, Ding, Ding, and
  Shi}]{wang-etal-2020-wmt20tencent}
Longyue Wang, Zhaopeng Tu, Xing Wang, Li~Ding, Liang Ding, and Shuming Shi.
  2020.
\newblock \href {https://aclanthology.org/2020.wmt-1.60} {Tencent {AI} lab
  machine translation systems for {WMT}20 chat translation task}.
\newblock In \emph{Proceedings of the Fifth Conference on Machine Translation},
  pages 483--491, Online. Association for Computational Linguistics.

\bibitem[{Zhang et~al.(2021)Zhang, Titov, Haddow, and
  Sennrich}]{zhang-etal-2021-context}
Biao Zhang, Ivan Titov, Barry Haddow, and Rico Sennrich. 2021.
\newblock \href {https://doi.org/10.18653/v1/2021.acl-long.200} {Beyond
  sentence-level end-to-end speech translation: Context helps}.
\newblock In \emph{Proceedings of the 59th Annual Meeting of the Association
  for Computational Linguistics and the 11th International Joint Conference on
  Natural Language Processing (Volume 1: Long Papers)}, pages 2566--2578,
  Online. Association for Computational Linguistics.

\end{thebibliography}
\bibliographystyle{acl_natbib}

\cleardoublepage

\appendix

\section{Crowdsourcing Details}
\subsection{Crowdsourcing Instructions Given to the Workers} \label{sec:crowdsourcing-instruction}

\begin{figure*}[htbp]
    \centering
	\subfigure[English]{
		\includegraphics[clip, width=0.45\linewidth]{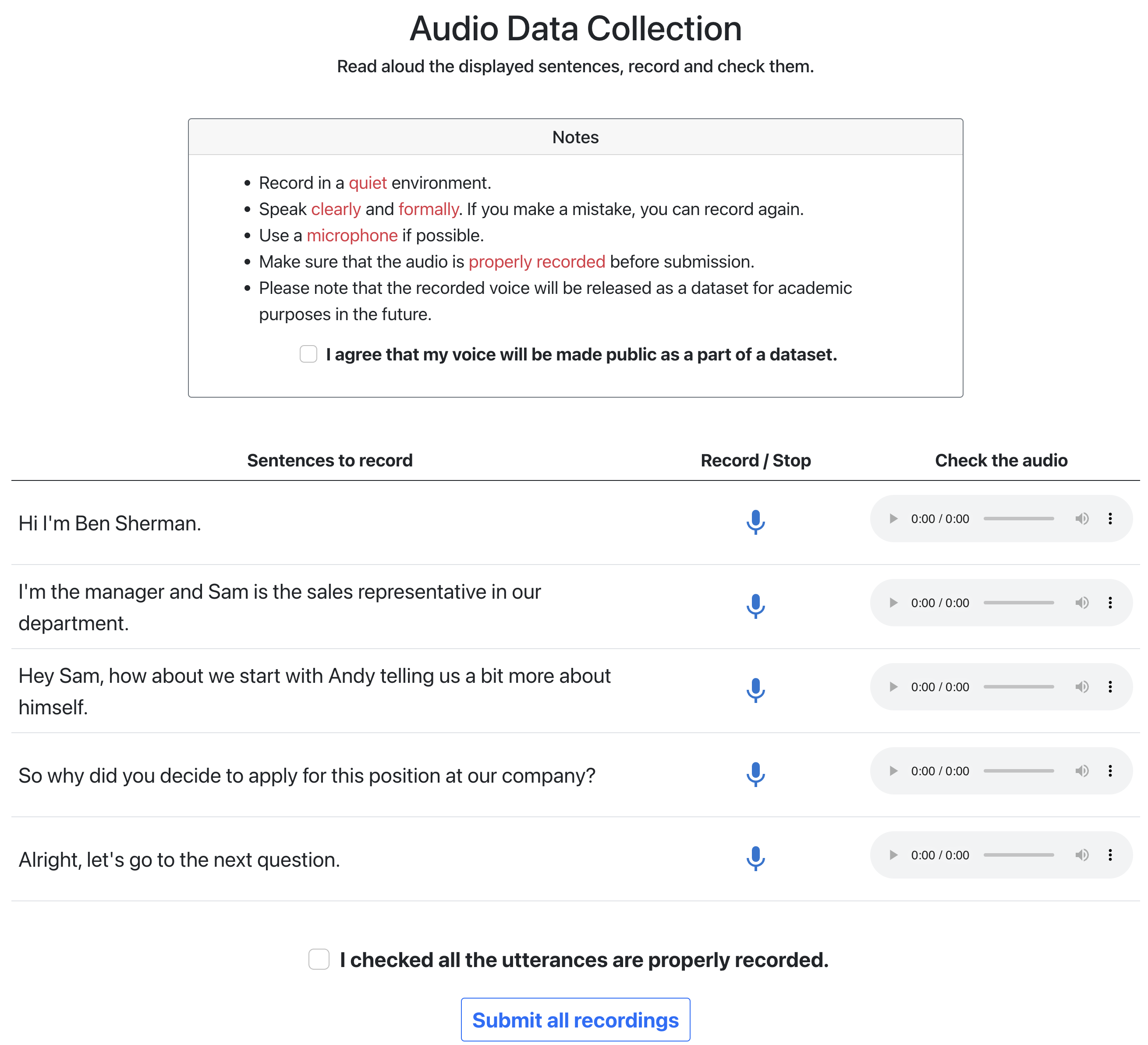}
		\label{subfig:en-crowd}
	}
	\hfill
    \subfigure[Japanese]{
		\includegraphics[clip, width=0.45\linewidth]{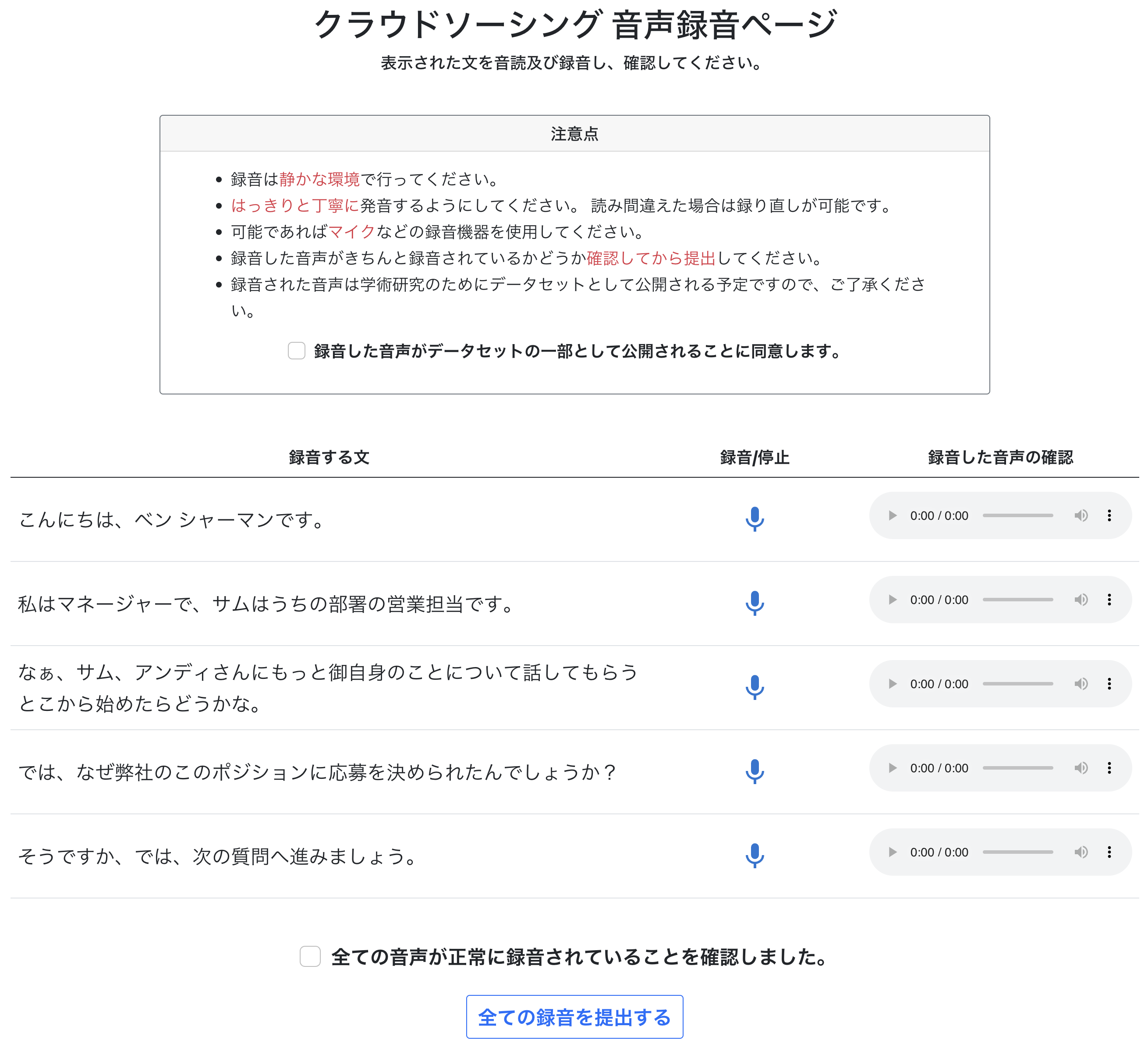}
		\label{subfig:ja-crowd}
	}
	\caption{Crowdsourcing interface used to record audio. The upper part shows the instructions given to the workers.}
	\label{fig:instruction}
\end{figure*}

Figure \ref{fig:instruction} shows the instructions given to the crowdsourcing workers and the interface used to record audio.
We asked the workers to speak clearly and formally and to check that the audio was properly recorded.
With the interface, we made sure that the workers agreed that their voices would be released and that the utterances were properly recorded.

\subsection{Crowdsourcing Payment}
The crowdsourcing tasks were divided according to the number of utterances to record.
The authors performed preliminary crowdsourcing tasks and estimated how long the tasks would take for each case.
We paid the workers according to the estimated time and predefined wage per hour determined for each country.

\section{Statistics of the SpeechBSD Dataset} \label{sec:appendix-speechbsd-statistics}
\begin{figure*}[htbp]
    \centering
	\subfigure[English]{
		\includegraphics[clip, width=0.5\linewidth]{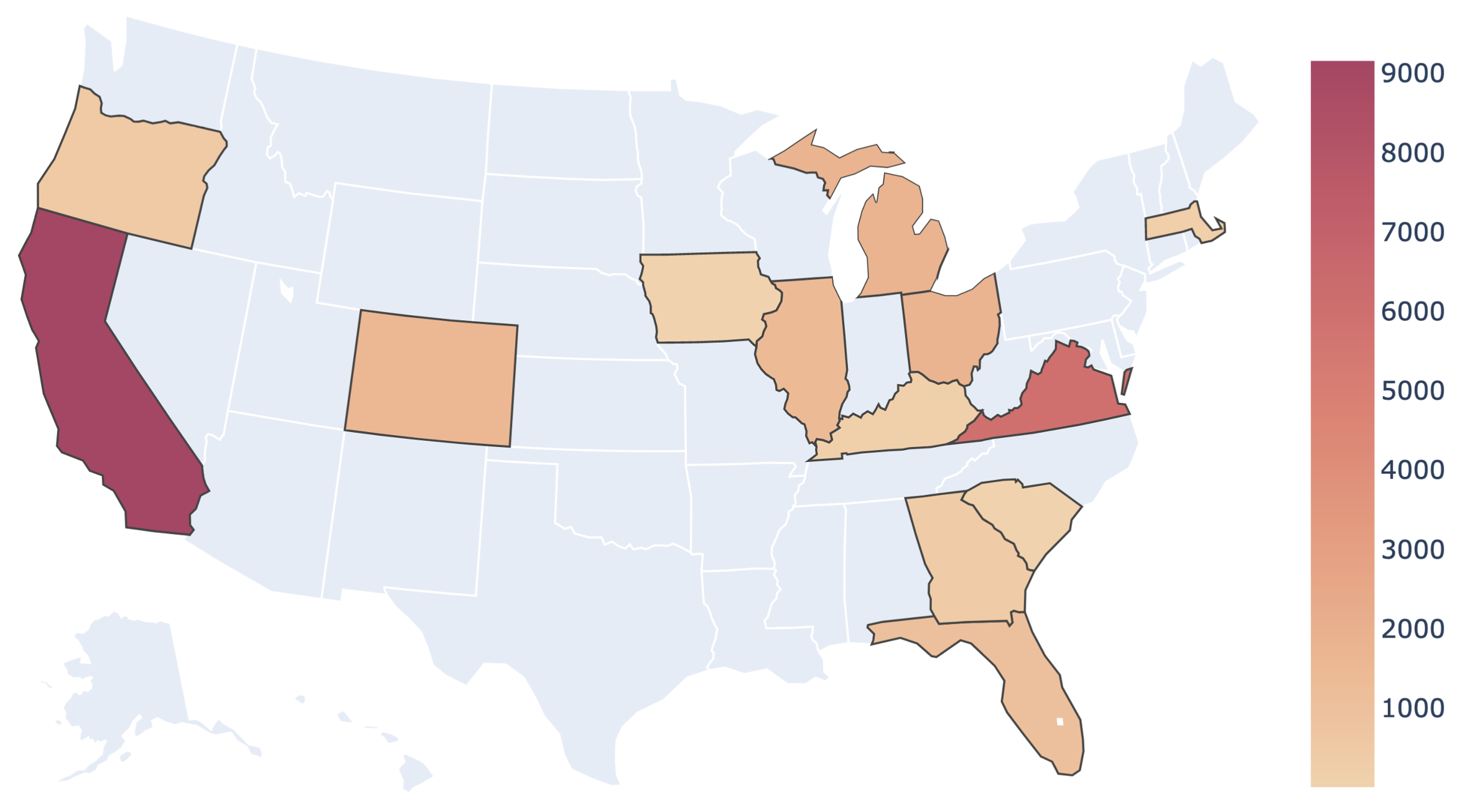}
		\label{subfig:en-dist}
	}
	\hfill
    \subfigure[Japanese]{
		\includegraphics[clip, width=0.4\linewidth]{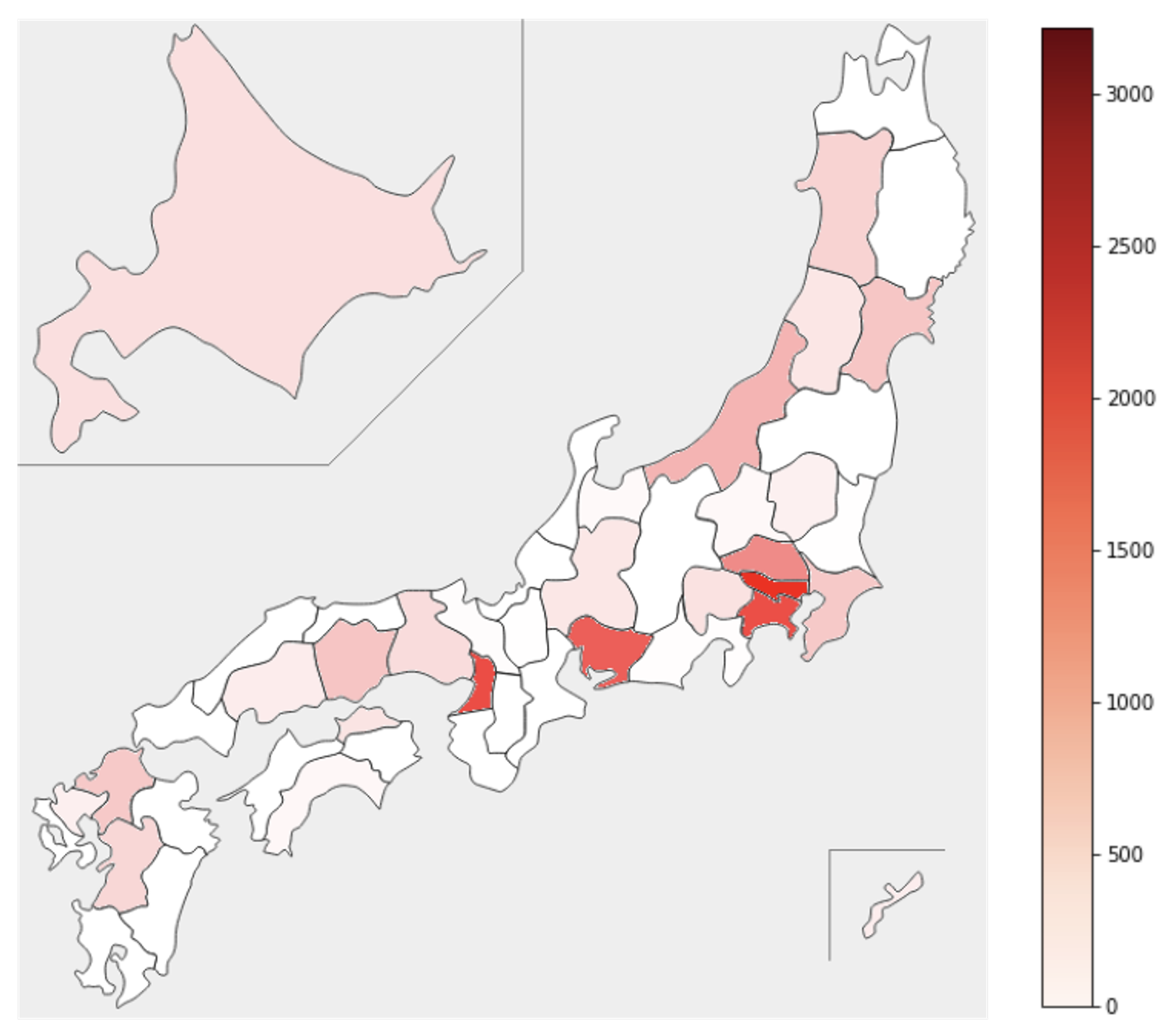}
		\label{subfig:ja-dist}
	}
	\caption{Homeplace distribution of the speakers of the SpeechBSD dataset by the number of utterances.}
	\label{fig:distribution}
\end{figure*}

Table \ref{tab:speechbsd} shows the statistics of the SpeechBSD dataset.
Figure \ref{fig:distribution} shows the homeplace distribution of the speakers of the SpeechBSD dataset.
The Japanese one (\ref{subfig:ja-dist}) roughly reflects Japan's demographics (concentrated around Tokyo, Osaka, and Nagoya), whereas the English one (\ref{subfig:en-dist}) is more biased (concentrated too much on California and Virginia).
We believe these biases are caused by the differences in the crowdsourcing platforms used.

\begin{table*}[t]
    \centering   
    \begin{tabular}{@{}crrr@{}}
    \toprule
              & Train & Dev. & Test \\ \midrule
    \# of scenarios & 670    & 69    & 69 \\
    \# of sentences & 20,000 & 2,051 & 2,120 \\
    English speech (h) & 20.1 & 2.1 & 2.1 \\
    Japanese speech (h) & 25.3 & 2.7 & 2.7 \\
    English gender (M / F \%) & 47.2 / 52.8 & 50.1 / 49.9 & 44.4 / 55.6 \\
    Japanese gender (M / F \%) & 68.0 / 32.0 & 62.3 / 37.7 & 69.0 / 31.0 \\ \bottomrule
    \end{tabular}
    \caption{Statistics of the SpeechBSD dataset. The number of sentences is the same as the number of utterances in this dataset as it originally was in the BSD corpus.}
    \label{tab:speechbsd}
\end{table*}

\section{Formal Definition of Context} \label{sec:formal-definition}

Here, we formally formulate monolingual, bilingual, and constrained contexts introduced in Section \ref{sec:context-sdt}.

For simplicity, we consider the case where $M = 2$ and $m = n$ (i.e., speaker $S^i$ speaks in language $L^i$ ($i = 1,2$)).
In addition, we suppose the speakers speak interchangeably, and speaker $S^1$ starts the conversation.\footnote{In the experiments, consecutive utterances by the same speaker are treated as separate utterances. If there are more than three speakers, we number speakers in the order of appearance and regard speakers with the same parity speak in the same language.}
In other words, defining a map $L: U_t \mapsto L^i$,
\[
    {}^\forall U \in \{ U_t \mid t \equiv i \, (\bmod 2) \}, \quad L(U) = L^i.
\]

The monolingual context is composed of previous utterances in a single language.
In other words, monolingual context text of utterance $U_t$ in language $L^i$ is
\[
    \bm{Y}^i_{<t} = \{ Y_\tau^i \, | \, \tau < t \}.
\]
For example in Figure \ref{fig:sdt}, when translating the third utterance $U_3$ from Japanese to English, the monolingual context of the source side is ``\begin{CJK}{UTF8}{ipxm}彼は良い考えだと言ってました。あなたはどう思いますか?\end{CJK}'', and that of the target side is ``He said it's a good idea. What do you think?''
Using this formulation, we can formally define the training objective of Equation \ref{eq:monocontext-objective}.
During inference, for the source language of the current utterance, ASR transcripts are used, and for the target language of the current utterance, the translations of ASR transcripts are used to compose context. During training, the corresponding gold text is used.

The Bilingual context is composed of transcripts of the two languages. ASR transcripts are used during inference, and gold transcripts are used for training.
The bilingual context of utterance $U_t$ is
$
    \bm{Y}_{<t} = \tilde{\bm{Y}}^1_{<t} \cup \tilde{\bm{Y}}^2_{<t},
$
where
\[
\tilde{\bm{Y}}^i_{<t} = \{ Y_\tau^i \mid \tau < t \, \land \, \tau \equiv i \, (\bmod 2) \}.
\]
For example in Figure \ref{fig:sdt}, when translating the third utterance $U_3$ from Japanese to English, the bilingual context of the source side is ``\begin{CJK}{UTF8}{ipxm}彼は良い考えだと言ってました。\end{CJK}What do you think about it?'', and that of the target side is ``He said it's a good idea. \begin{CJK}{UTF8}{ipxm}あなたはどう思いますか?\end{CJK}''

For bilingual context experiments, the MT system has to be able to handle two translation directions.
Let the translation of $\bm{Y}_{<t}$ be
$
\overline{\bm{Y}_{<t}} = \overline{\tilde{\bm{Y}}^1_{<t}} \cup \overline{\tilde{\bm{Y}}^2_{<t}},
$
where
\begin{align*}
\overline{\tilde{\bm{Y}}^i_{<t}} = \{ Y_\tau^j \mid \tau < t \, \land \, \tau \equiv i \, (\bmod 2) \}, \\
(i,j) = (1,2), (2,1).
\end{align*}
$\overline{Y_t}$ is $Y_t^2$ when $L(U_t) = L^1$ and $Y_t^1$ when $L(U_t) = L^2$.
By setting $\bm{Y}_{<t}$ as source side context and target side context as $\overline{\bm{Y}_{<t}}$, we can formally define the training objective of Equation \ref{eq:bicontext-objective}.

In practice, we consider context width $c$ for context $\bm{U}_{<t} = \{ U_\tau \, | \, \tau < t \}$ because the maximum length the MT models can handle is limited.
The constrained context of utterance $U_t$ with context width $c$ is
$$
    \bm{U}_{<t} = \{ U_\tau \, | \, \tau = t - 1, \cdots, t - c \, \land \, \tau > 0\}.
$$

\section{Experimental Settings}
\subsection{ASR} \label{sec:asr-settings}
Whisper is a Transformer-based model that uses $80$-channel log-Mel spectrograms converted from audio sampled with $16,000$ Hz as input.
As it is trained with $680,000$ hours of data in various domains the model is robust enough to be able to work without any finetuning.
We used the byte-level BPE vocabulary (size $50,257$) of the pretrained model.
We assumed the language of the utterances was given beforehand and fed the language tag to the model as a prefix token.
We evaluated the development set of the SpeechBSD dataset using the base, small, medium, and large models with either greedy decoding or beam search decoding with beam size $5$.
We observed that the medium model with greedy decoding performed the best for both English and Japanese, which are the settings used for further experiments.

\subsection{MT} \label{sec:mt-settings}
We used mBART trained with 25 languages for the experiments.
BPE vocabulary of size $25,001$ was used.
As a preprocessing step, BPE was applied to all utterances with the sentencepiece \citep{kudo-richardson-2018-sentencepiece} toolkit.
Fairseq \citep{ott-etal-2019-fairseq} was used for training and inference.
The same hyperparameters as in \citet{liu-etal-2020-mbart} were used, except that the training epochs were determined according to early stopping with patience $10$ on validation loss.
We did not use different random seeds for the experiments because \citet{liu-etal-2020-mbart} reported that the finetuning process was stable with different seeds.
When evaluating the model, the averaged weights of the last $10$ checkpoints were used.
The SacreBLEU signatures were \texttt{nrefs:1|case:mixed|eff:no|tok:ja-mecab-\\0.996-IPA|smooth:exp|version:2.0.0} for En--Ja and \texttt{nrefs:1|case:mixed|eff:no|tok:13a|\\smooth:exp|version:2.0.0} for Ja--En.
We conducted significance tests with paired approximate randomization \citep{riezler-maxwell-2005-pitfalls} with $10,000$ approximate randomization trials and a $p$-value threshold of $0.05$ to compare the BLEU scores of ``without context'' with the others, and ``monolingual context'' with ``bilingual context.''

For bilingual context MT experiments, in order to match the finetuning style of mBART, language tags like \texttt{ja\_XX} or \texttt{en\_XX} have to be appended at the last of each translation unit.
However, in bilingual context settings, both the source and the target side contain both languages, which does not comply with the finetuning style described in the original mBART paper \citep{liu-etal-2020-mbart}.
We conducted two kinds of experiments, appending \texttt{ja\_XX} to the input and \texttt{en\_XX} to the output and the other way around.
The statistical significance test showed that they were not significantly different.
We report the results of the systems where the language pair of the utterance to be translated matches the language pair specified by the appended language tags.

\begin{figure}[htbp]
    \centering
    \includegraphics[width=\linewidth]{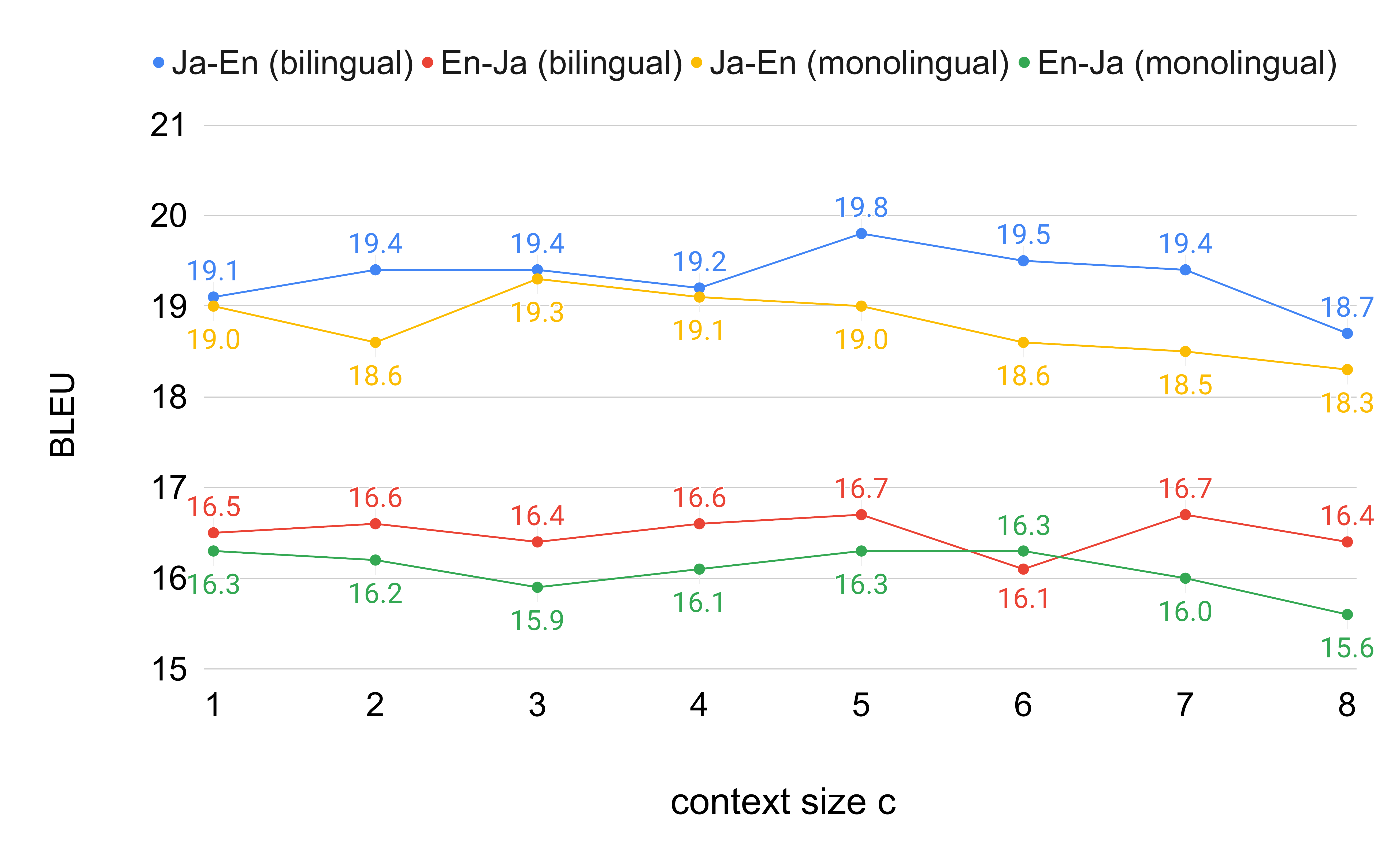}
    \caption{BLEU score on the development set when changing the context size $c$.}
    \label{fig:vary-context-size}
\end{figure}

As to the context size $c$, we changed it from $1$ to $8$ in the bilingual context setting and evaluated the models with BLEU score on the validation set.
The results are shown in Figure \ref{fig:vary-context-size}.
In the bilingual context setting, $5$ was the best for both En--Ja and Ja--En.
For the monolingual context setting, $5$ and $6$ were the best for En--Ja and $3$ for Ja--En. 
The difference between setting $3$ and $5$ as context width did not show a statistically significant difference in the BLEU scores for Ja--En.
Therefore, for a consistent comparison, we reported the results on the test set with $c = 5$ in Table \ref{tab:mt-and-cascade}.

We used $4$ Tesla V100 or Titan RTX GPUs for the experiments.
The total computation hours, including hyperparameter searching, were $278$ hours.

\section{Example Sentences from Manual Evaluation} \label{sec:manual-eval-example}

Table \ref{tab:manual-eval-example} shows examples from manual evaluation described in Section \ref{sec:manual-eval}.
In the first example, it is observed that the zero pronoun (\textit{She}) is predicted correctly when monolingual or bilingual context is used in both MT and cascade ST experiments.
In the second example, the zero pronoun (\textit{They}) could not be correctly predicted by any system.

\begin{table*}[t]
    \centering

    \begin{tabular}{@{}ccl@{}}
    \toprule
     & Context & \\ \midrule
     Ja reference & - & \begin{CJK}{UTF8}{ipxm}もう諦めて、仕事なら仕方ないわねって。\end{CJK} \\
     En reference & - & She's given up and just says it can't be helped if it's work. \\ \midrule
    \multirow{3}{*}{MT} & Without & I just gave up and said I can't do it if it's my job. \\
     & Monolingual & She just gave up and said it's okay if it's the job. \\
     & Bilingual & She's giving up on it and says if it's work then it's all right. \\ \midrule
    \multirow{3}{*}{Cascade ST} & Without & If I just gave up and gave up my job, I can't do anything about it. \\
     & Monolingual & She said if I just gave up and gave up on the job then it should be fine. \\
     & Bilingual & She said if I just give up and give up on the job then it's all right. \\ \bottomrule
    \end{tabular}

    \vspace{1em}
    (a) An example where the ``monolingual'' and ``bilingual'' context predictions were better than the ``without context'' one. In this scenario, Patrick complains to Gary that he does not want to go to his company's drinking party. Gary asks what Patrick's wife thinks about it, and this is Patrick's response. The pronoun \textit{She} is omitted in the Japanese utterance. Word-by-word translation of the Japanese utterance with omitted words is: ``\begin{CJK}{UTF8}{ipxm}(彼女は)--she / もう--already / 諦めて--give up / それが--it's / 仕事--work / なら--if / (それは)--it / 仕方ない--can't be helped / わね--I think / って(言ってる)--says \end{CJK}.''
    
    \vspace{2em}
    
    \begin{tabular}{@{}ccl@{}}
    \toprule
     & Context & \\ \midrule
     Ja reference & - & \begin{CJK}{UTF8}{ipxm}いつ在庫が入るか、でしょう?\end{CJK} \\
     En reference & - & They all want to know when it will be restocked, don't they? \\ \midrule
    \multirow{3}{*}{MT} & Without & When will the inventory start? \\
     & Monolingual & So when will the inventory be available? \\
     & Bilingual & I wonder when it will be in stock? \\ \midrule
    \multirow{3}{*}{Cascade ST} & Without & When will the inventory arrive? \\
     & Monolingual & I wonder when it will be in stock. \\
     & Bilingual & I wonder when it will be in stock. \\ \bottomrule
    \end{tabular}
    
    \vspace{1em}
    (b) An example where all systems failed to predict the correct pronoun. In this scenario, Mr. Ogawa and Ms. Pace are talking about their company's stock of a product. The previous utterances by Mr. Ogawa are, ``We have 28 backorders for this product. I have been receiving many inquiries from the customers lately.'' This is the subsequent Ms. Pace's response. The pronoun \textit{They} is omitted in the Japanese utterance. Word-by-word translation of the Japanese utterance with omitted words is: ``\begin{CJK}{UTF8}{ipxm}(彼らは)--they / いつ--when / 在庫--stock / が入るか--becomes avaiable / (を聞くの)--ask  / でしょう--don't they\end{CJK}.'' The translation is difficult because the word corresponding to ``ask'' is also omitted.
    
    \caption{Examples from manual evaluation of Ja--En translations focusing on zero pronouns.}
    \label{tab:manual-eval-example}
\end{table*}

\end{document}